\documentclass[a4paper,twoside]{article}

\usepackage{epsfig}
\usepackage{graphicx}
\usepackage{wrapfig}
\usepackage{xcolor}
\usepackage{color,colortbl}
\usepackage{subcaption}
\usepackage{calc}
\usepackage{amssymb}
\usepackage{amstext}
\usepackage{amsmath}
\usepackage{amsthm}
\usepackage{multicol}
\usepackage{pslatex}
\usepackage{apalike}
\usepackage{SCITEPRESS}     

\definecolor{Gray}{gray}{0.9}
\begin{document}

\title{Deep Learning for Posture Control Nonlinear Model System and Noise Identification.}

\author{\authorname{Vittorio Lippi\sup{1}, Thomas Mergner\sup{2},Christoph Maurer\sup{2}}
\affiliation{\sup{1}Technical University Berlin, Control Systems, Berlin, Germany}
\affiliation{\sup{2}Neurological University Clinic, University of Freiburg, Freiburg im Breisgau, Germany}
\email{vittorio.lippi@tu-berlin.de, thomas.mergner@uniklinik-freiburg.de, christoph.maurer@uniklinik-freiburg.de}
}


\abstract{In this work we present a system identification procedure based on Convolutional Neural Networks (CNN) for human posture control models. A usual approach to the study of human posture control consists in the identification of parameters for a control system. In this context, linear models are particularly popular due to the relative simplicity in identifying the required parameters and to analyze the results. Nonlinear models, conversely, are required to predict the real behavior exhibited by human subjects and hence it is desirable to use them in posture control analysis. The use of CNN aims to overcome the heavy computational requirement for the identification of nonlinear models, in order to make the analysis of experimental data less time consuming and, in perspective, to make such analysis feasible in the context of clinical tests. Some potential implications of the method for humanoid robotics are also discussed.}

\onecolumn \maketitle \normalsize \setcounter{footnote}{0} \vfill

\section{\uppercase{Introduction}}
\label{sec:introduction}

Mathematical models of human posture control are used in the analysis of experiments as well as in the control of humanoid robots. For this reason system identification techniques have been developed for the identification of human balance as dynamic system with feedback control \cite{vanderKooij2007,van2005comparison,van2006disentangling,goodworth2018identifying,mergner2010,engelhart2014impaired,pasma2014impaired,jeka2010dynamics,boonstra2014balance}. Most of the studies performed on human posture control exploit linear models such as the \textit{independent channel} model \cite{peterka2002sensorimotor}, and in general assume a linear and time invariant behavior for human posture control \cite{engelhart2016comparison}. Linear models have the advantage of being simple to analyze and relatively easy be fit on the data. However experiments reveal that human posture control exhibits nonlinearities such as dead-bands and gain-non-linearity. Nonlinear models are more complex to be fit on human data and, in the general case expensive iterative procedures should be used. In this work we propose a deep learning system to identify the parameters of a nonlinear bio-inspired posture control system, the DEC (\textit{Disturbance Estimation and Compensation}). The obtained set of parameters represents a concise and expressive representation of  the outcome of a posture control experiment that can be used for scientific studies and as a basis of future diagnostic tools for clinicians. The proposed technique is based on \textit{Convolutional Neural Networks}, CNN. Such deep learning system, has been recently applied with promising result to human movement analysis, e.g. in \cite{icinco19}, but so far it has not been a tool typically used in posture control analysis.
\section{\uppercase{Methods}}
\subsection{Posture Control Scenario: Support Surface tilt}
The scenario considered here models a human (or humanoid) balancing on a tilting support surface. The support surface tilt $\alpha_{FS}$ represents the input of the system and it is the same for all the simulations. The profile of the tilt of the support surface is the \textit{pseudo-random ternary sequence}, PRTS, shown in Fig. 1 (C). Such stimulus is used in human experiments because, thanks to its pseudo-random nature, it is not predictable for the subject\cite{peterka2002sensorimotor}. Furthermore it is composed by a sequence of velocity steps suitable to excite the dynamics of the system over several frequencies. The output of the system is the sway of the COM $\alpha_{BS}$
\subsection{Human and Humanoid Posture Control: The DEC model}
\begin{figure}[t]
	\centering
		\includegraphics[width=1.00\columnwidth]{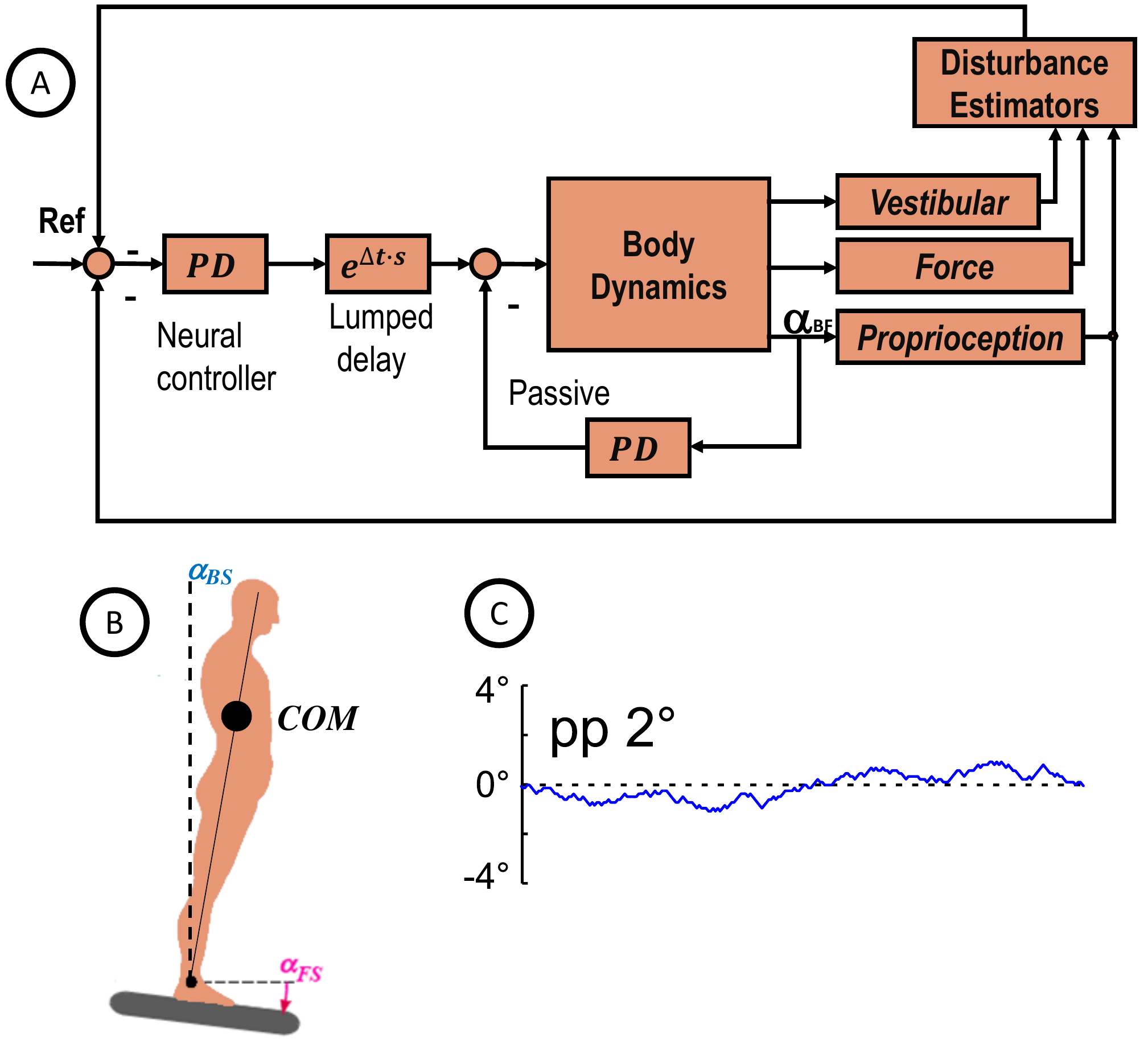} \\ 
	\caption{A scheme of a controller based on the DEC concept. (A) the sensory inputs are used to reconstruct the physical disturbances acting on the body. Such disturbances are compensated feeding them in the form of an \textit{angle equivalent} as input to the servo controller. (B) The single inverted pendulum model used to simulate human posture control. The kinematics are described by the COM sway angle $\alpha_{BS}$ and by the support surface tilt $\alpha_{FS}$. (C) The \textit{Pseudo-Random Ternary Signal}, PRTS, used as reference for the support surface tilt.} 
	\label{DEC}
\end{figure}

The DEC concept provides a a model of the human postural control mechanisms \cite{mergner2010,lippi2017human}. This approach has been applied to multiple DoF robots \cite{lippi2017human,lippi2017human,zebenay2015human,ott2016good,lippi2018prediction,Hettich2013,hettich2015human}. In this work the implications of a modular control architecture will not be covered because the proposed single inverted pendulum (SIP) model consists only of one control module. A general scheme describing the DEC control is shown in Fig. \ref{DEC}.
In general a posture control system based on the DEC concept is implemented as: 
(1) A servo loop, implemented as a PD controller (neural controller in Fig. \ref{DEC}). In the presented 1DoF case the controlled variable consists in the body center of mass with sway respect to the gravitational vertical passing through the ankle joint $\alpha_{BS}$, where $BS$ stands for \textit{Body in Space}. (2) The disturbance estimations are, in general, identifying support surface rotation and \textit{translation}, external contact forces  and field forces such as gravity . The sensory channels are shown in Fig. \ref{DEC} as \textit{Vestibular}, \textit{Proprioceptive}, and \textit{Force}. The disturbance estimates are fed into the servo so that the joint torque on-line compensates for the disturbances while executing the desired movements. The \textit{lumped delay} in Fig. \ref{DEC} represents  all the delay effects that in humans (but also in real world humanoids) are distributed \cite{antritter2014stability,G.Hettich2014}.
The model used in this work considers gravity and support surface tilt as disturbances. Specifically the estimators are defined as follows:\\
\underline{Gravity estimator}
\begin{equation}
	T_G = G_g \alpha_{BS}^{vest}
\end{equation}
where $G_g$ is a gain associated with the estimator. In the framework of the DEC control the disturbances are represented by an angle equivalent, i.e. the body lean that would produce, in a linear approximation, the disturbance torque as gravitational torque. This makes all the values that are summed to obtain the input of the neural controller (error and disturbances) expressed in radians. In the specific case of the gravitational torque the equivalent angle is the body lean $\alpha_{BS}^{vest}$.  With an ideal $G_g$ of $1$ and a proportional gain $K_p = mgh$, where $m$ is body mass, $g$ gravity acceleration and $h$ is the height of the COM respect to the ankle joint, the gravity would be exactly compensated. When fitting the model to human behavior the gravity appears to be slightly under-compensated \cite{G.Hettich2014,asslander2015visual}. In this work $G_g$ will be set to 1, and hence the gravity compensation gain will be determined by $K_p$.
The signal $\alpha_{BS}^{vest}$ comes from the vestibular system and it is affected by a \textit{red noise} $\nu(N_V)$ with frequency power density $N_v^2/f$, where $N_v$ is a parameter of the system.\\
\underline{Support surface tilt estimator}
\begin{equation}
	\alpha_{FS}=\int_0^t f_{\theta} \left( \frac{d}{dt} \alpha_{BS}^{vest} + \frac{d}{dt} \alpha_{BF}^{prop} \right)
	\label{fs}
\end{equation}
where $\alpha_{BF}^{prop}$ is the ankle joint angle signal from proprioception. $BF$ stands for \textit{Body-to-Foot}. In some implementations of the DEC concept the integral in eq. \ref{fs} is implemented as a leaky integrator \cite{lippi2017human}, in this work it is set to zero at the beginning of the simulation. The function $f_{\theta}$ is a dead-band threshold defined as
\begin{equation}
	f_{\theta}(\alpha) = \left\{ 	
	\begin{array}{llc}
   		\alpha + \theta & if & \alpha< -\theta \\
			0 & if & -\theta< \alpha< \theta \\
			\alpha - \theta & if & \alpha> \theta \\
 	\end{array}
	\right.
\end{equation}
The threshold function is added to reproduce the behavior observed in humans \cite{T.Mergner2009,Mergner2003}. The reconstructed body-in-space variable used for the servo controller is then
\begin{equation}
	\alpha_{BS}^{servo}=\alpha_{FS}-\alpha_{BF}^{prop}
\end{equation}
affected by the non-linearity introduced by $f_{\theta}$. Considering that in the present scenario the simulated agent aims to maintain the upright stance, i.e. the reference signal is zero, the total torque commanded by the servo controller is:
\begin{equation}
	\tau_{active}=-e^{-s\Delta } (K_p+s K_d)\left( T_g+ \alpha_{BS}^{servo}\right)
	\end{equation}
where $K_d$ is the derivative coefficient for the PD controller (for the sake of brevity in this equation and the following the derivatives, the integrators and the delay are represented using Laplace transform variable $s$ so that they can be conveniently expressed as a multiplicative operator, although the rest of the formula refers to operations in time domain). $\Delta$ is the lump delay. Notice that the derivative component is acting also on gravity compensation, representing a sort of anticipation of the disturbance. There is also a passive torque acting on the ankle joint defined as:
\begin{equation}
	\tau_{passive}=-(K_p^{pass}+s K_d^{pass})\left(\alpha_{BF}^{prop}\right)
	\end{equation}

In order to show the role of all the parameters (listed in Table \ref{tab:parameters}, here highlighted in blue) the total torque can be written as:
\begin{equation}
\begin{array}{l}
	\tau_{ankle}= \tau_{active}+\tau_{passive} \\ \\
	=-e^{-s\color{blue}{\Delta} }({\color{blue}{K_p}}+s {\color{blue}{K_d}}) \\ \\
	\left(\alpha_{BS}^{vest} + \alpha_{FS}	-\frac{1}{s} f_{\color{blue}{\theta}} \left( s (\alpha_{BS}^{vest} +\nu({\color{blue}{N_v}})) + s \alpha_{BF}^{prop} \right) \right) \\ \\
	-({\color{blue}{K_p^{pass}}}+s {\color{blue}{K_d^{pass}}})\left(\alpha_{BF}^{prop} \right)
\end{array}
\end{equation}

\subsection{The Training set}
\begin{figure}[ht]
	\centering
		\includegraphics[width=1.00\columnwidth]{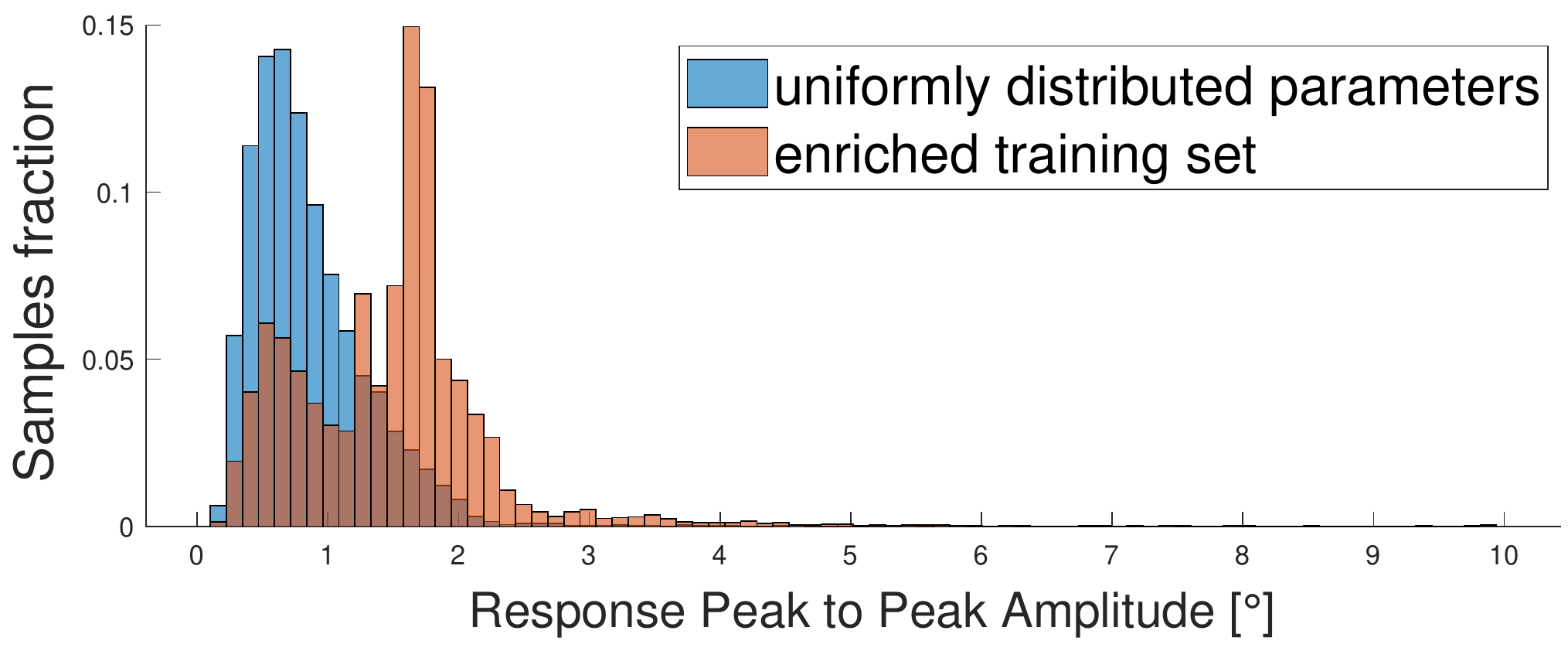}
	\caption{Peak-to-peak body sway amplitude distribution}
	\label{fig:HIST}
\end{figure}

The training and the validation set for the neural network have been generated with random parameters from uniform distributions (the range is shown in Table. \ref{tab:parameters}). A set of parameters is used as a sample only if the behavior it produces is stable: simulations with $\alpha_{BS}$ amplitude larger than $5^{\circ}$ are not considered realistic balancing scenarios and are discarded. Most of the stable simulations obtained with such random sampling were associated with a relative small COM sway, the amplitude distribution is shown in Fig. \ref{fig:HIST} in blue. In order to obtain a data-set including larger oscillations, representative of a \textit{relaxed} human behavior, the data-set was enriched with samples that were produced repeating the simulations with larger outputs ( $> 0.05 rad$) with parameters subject to a relatively small modification ($\approx 10\%$ of the range). The resulting enriched data-set is shown in Fig. \ref{fig:HIST} in orange. The performance of the neural network on the two sets was almost the same and hence only the enriched data-set is considered. The distribution of realistic human data-set is discussed more in detail in section \ref{sec:conclusion}. The resulting data-set included $12766$ samples. Half of the samples are used for training, half as validation set.

The neural network is trained to identify the simulation parameters on the basis of body sway profiles. The \textit{Target} for the training is represented by the vector of parameters, centered with respect to the mean normalized by the standard deviation (both computed on the training set). The \textit{Input} is  a convenient representation of the output. The simulation was performed with a fixed integration step of $1 ms$ and produced $12100$ $\alpha_{BS}$ samples with a resolution of $10 \; ms$. In order to adapt the signal to the convolutional network used the input was transformed into a two channel image, with the channels representing, respectively, the modulus and the phase of the FFT of the signal computed on non overlapping time windows. Empirical tests have shown that the best performance was achieved with a time window of $110$ samples resulting in a square $110 \times 110$ two-channel image (Fig. \ref{fig:Neuralarchitecture} above). 

\begin{table*}[t!]
\center
\renewcommand{\arraystretch}{2}
\begin{tabular}{|l|l|l|l|l|l|l|}
\hline
Parameter                        & Symbol         & min      & max       & mean     & std      & unit                            \\ \hline
\rowcolor{Gray}
Active proportional gain         & $K_p$          & 503.3943 & 1258.4857 & 811.2951 & 338.0956 & $\frac{N \cdot m}{rad}$         \\
Active derivative gain           & $K_d$          & 125.8486 & 377.5457  & 284.5640 & 122.7999 & $\frac{N \cdot m \cdot s}{rad}$ \\
\rowcolor{Gray}
Passive stiffness                & $K_{p_{pass}}$   & 62.9243  & 377.5457  & 312.2075 & 102.1054 & $\frac{N \cdot m}{rad}$         \\
Passive damping                  & $K_{d_{pass}}$   & 62.9243  & 188.7729  & 174.3144 & 68.5447  & $\frac{N \cdot m \cdot s}{rad}$ \\
\rowcolor{Gray}
Vestibular noise gain            & $N_V$          & 0        & 1.0000    & 0.4695   & 0.2928   & $1$                             \\
Foot rotation velocity threshold & $\theta_{vfs}$ & 0        & 0.0052    & 0.0003   & 0.0124   & $rad / s$                       \\ 
\rowcolor{Gray}
Lumped delay                     & $\Delta$       & 0        & 0.2400    & 0.1210   & 0.0672   & $s$                             \\ \hline
\end{tabular}
\caption{Simulation parameters with an overview of their distributions in the examples used as training and validation set. \textit{Min} and \textit{Max} represent the minimum and the maximum values of the uniform distributions used to generate the samples. \textit{Mean} and \textit{std} (standard deviation) are computed on the the selected simulations that resulted in a stable behavior (i.e. maximum $\alpha_{BS}$  oscillation under $5^{\circ}$) and included the \textit{enrichment} (see text).}
\label{tab:parameters}
\end{table*}

\subsection{The Neural Network}
\begin{figure}[htbp]
	\centering
		\includegraphics[width=1.00\columnwidth]{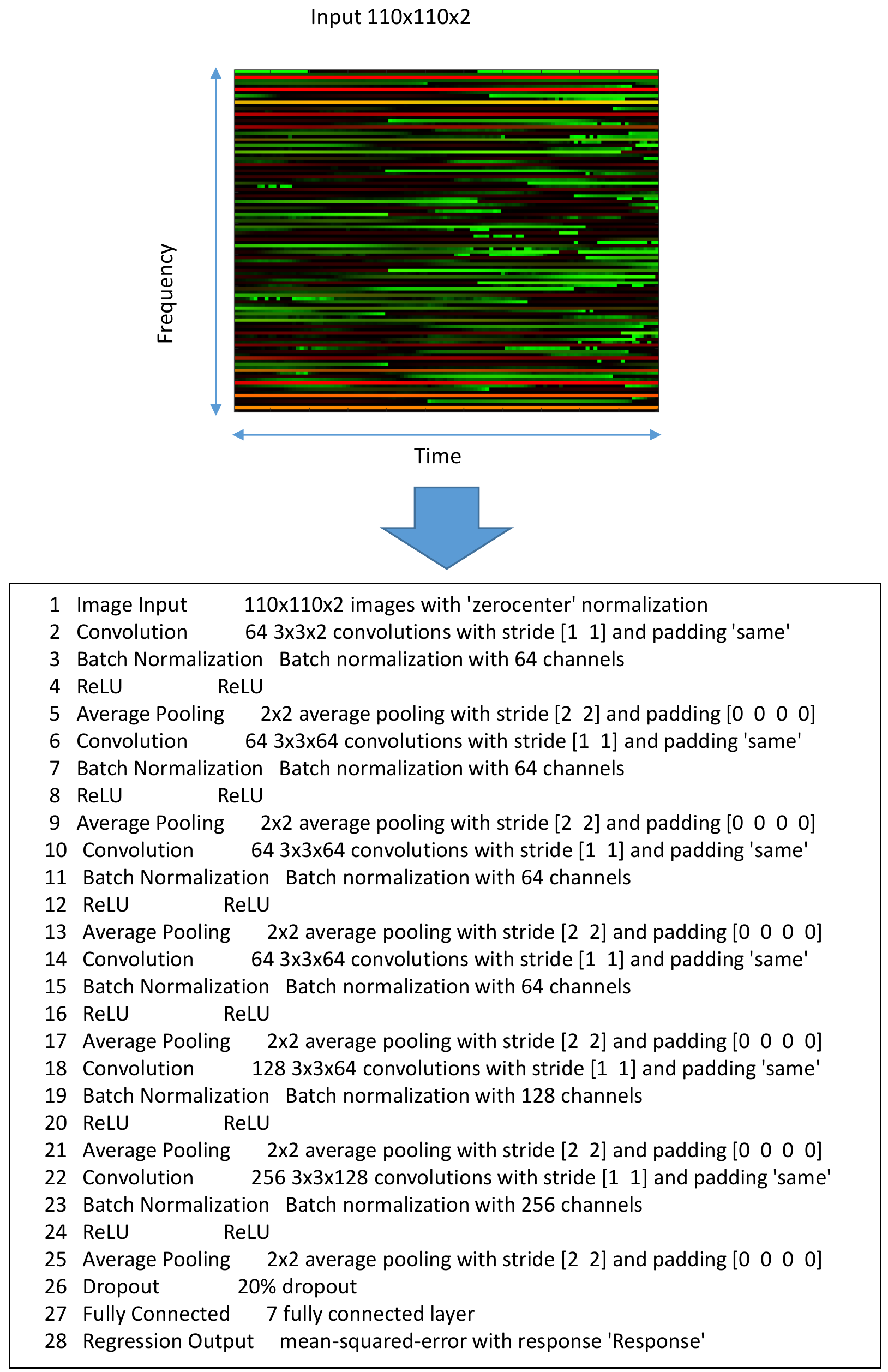}
	\caption{(Top) Example of the appearance of an input sample. The image has two channels, red and green, associated to the modulus the phase of the FFT of the body sway over a time window, respectively. (Bottom) Neural network architecture (below).}
	\label{fig:Neuralarchitecture}
\end{figure}

\begin{figure}
	\centering
		\includegraphics[width=1.00\columnwidth]{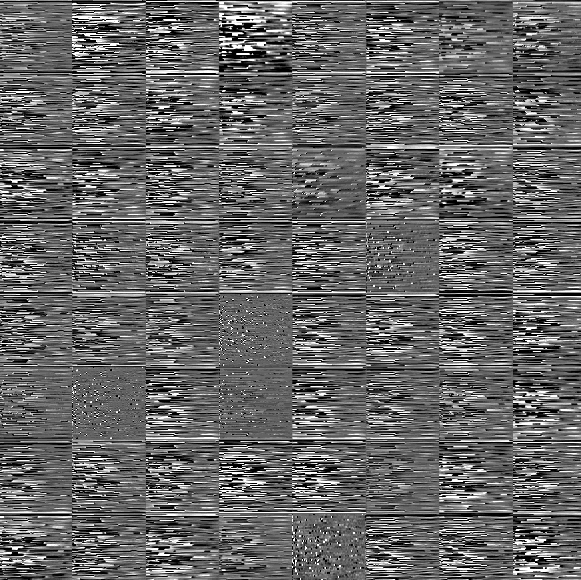}
	\caption{Graphic rendering of the first convolutional layer of the network. Each image in the $8 \times 8$ grid represent the response of one of the 64 filters to the input image. Lighter pixels are associate with a larger activation than darker pixels.}
	\label{fig:Activation}
\end{figure}

The neural network architecture is shown in Fig \ref{fig:Neuralarchitecture}, where the layers are listed.  The network has been implemented with Deep Learning Toolbox\texttrademark. Such network is not designed for 1-D convolution and hence the input is transformed into an image. The filters of convolutional layers apply the same weight to different parts of the input. The axis of the input image can be seen as \textit{time} and \textit{frequency}. This means that the convolutional filters allow the network to recognize pattern translated in time (horizontal) and in frequency (vertical). While the former invariance has the understandable meaning that the network is able to recognize the same movement patterns appearing at different times, as expected from an 1-D CNN applied to time series, the latter has no straightforward physical explanation. An example of the activation of the filters in the first layers is shown in Fig. \ref{fig:Activation}.

\begin{figure}[ht]
	\centering
		\includegraphics[width=1.00\columnwidth]{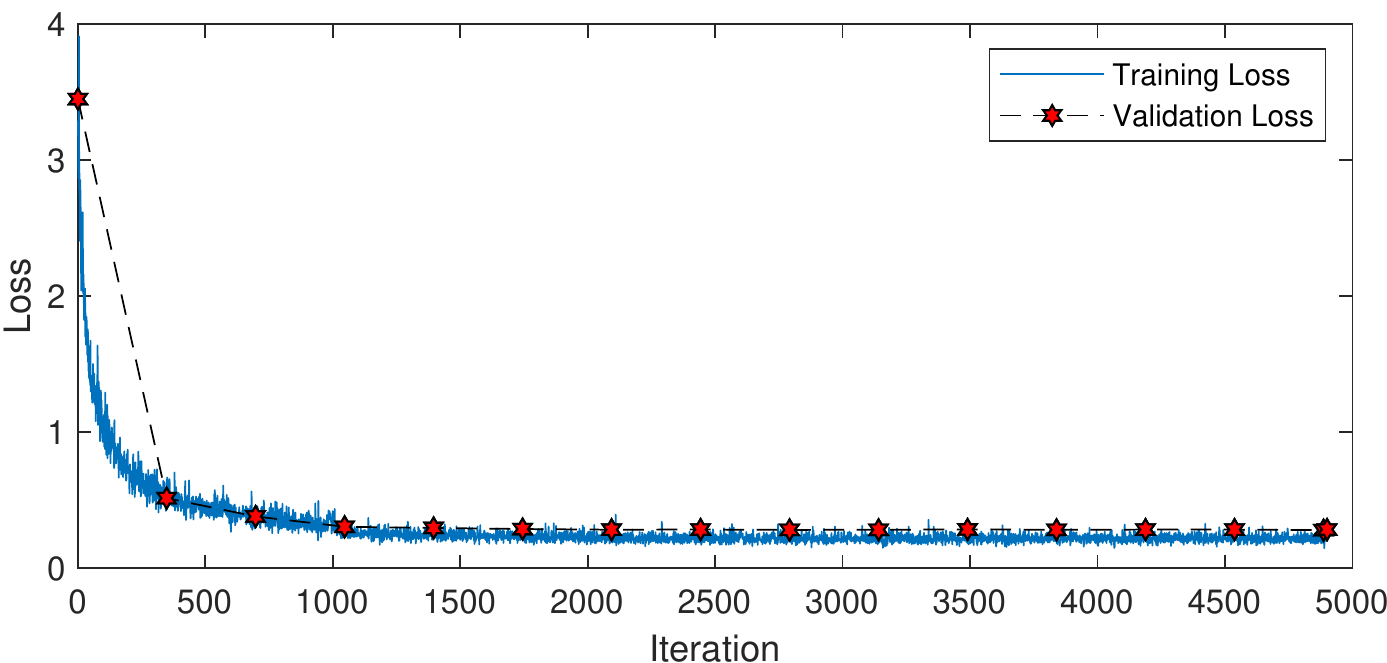}
	\caption{Training and validation loss through the iterations}
	\label{fig:TrainingLoss}
\end{figure}

The network has been trained using \textit{stochastic gradient descent with momentum} as policy. The training was set to a limit of 100 epochs. The data-set is divided in two equally sized subsets of $6383$ samples used as training set and validation set, respectively. The loss function is the \textit{Mean Squared Error} MSE, coherently for the regression task. The evolution of the MSE through the training iterations is shown in Fig. \ref{fig:TrainingLoss}.

\section{Results}
\subsection{Validation Set}
\begin{figure}[htbp]
	\centering
		\includegraphics[width=1.00\columnwidth]{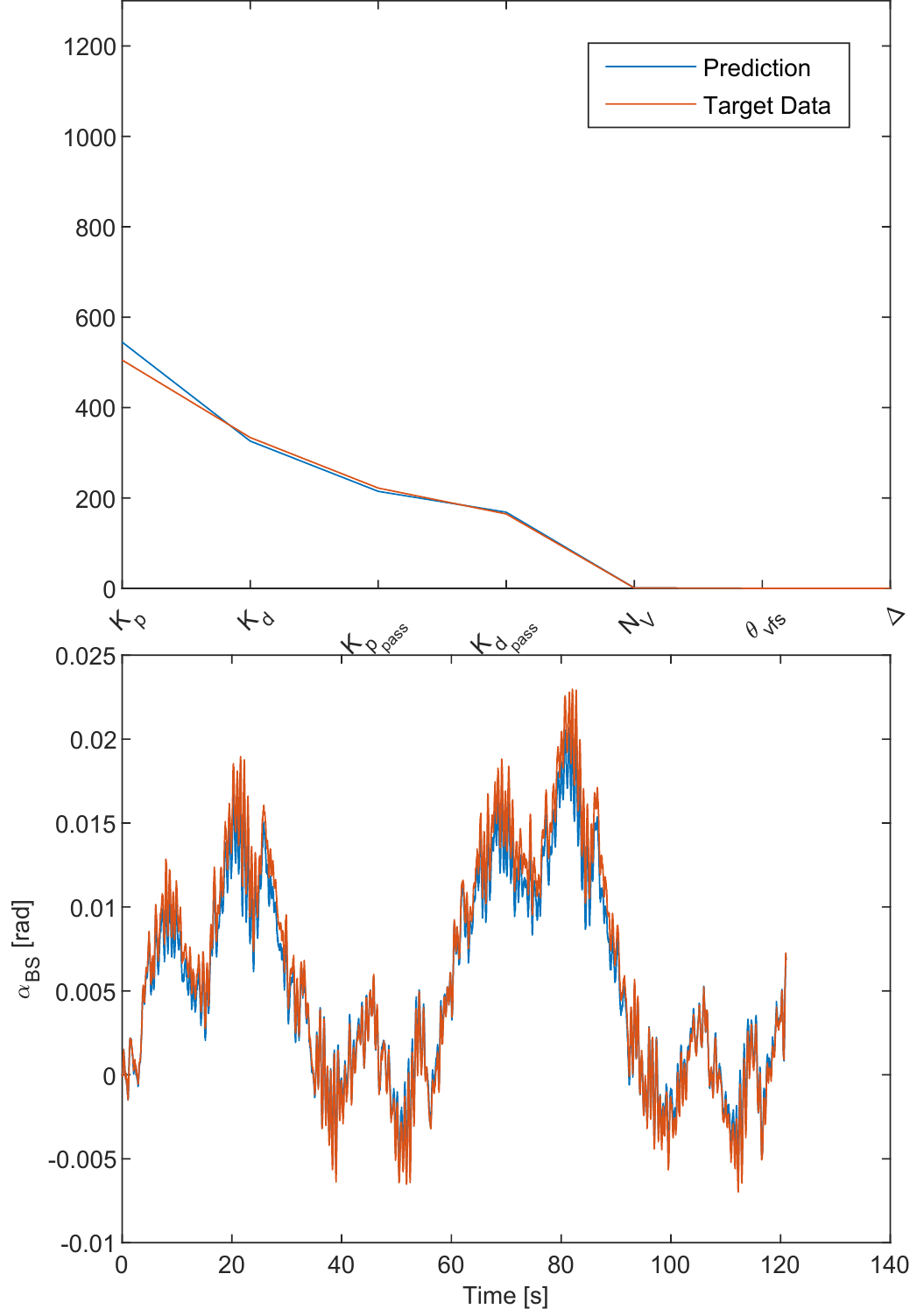}
		\caption{The CNN applied to a specific sample: (Top) network output and target sample, and (Bottom) the associated $\alpha_{BS}$.}
	\label{Res}
\end{figure}

The validation set loss, MSE, is $0.2851$. It is comparable to the one obtained for the training set which was $0.2250$. Figure \ref{fig:TrainingLoss} shows how the loss function is stable for several iteration on both training set and validation set. Thanks to the availability of a large enough number of samples the system is not showing signs of over-fitting. An example of the output obtained using a specific sample is shown in Fig. \ref{Res}.

\subsection{Double inverted pendulum case}
\begin{figure}[htbp]
	\centering
		\includegraphics[width=1.00\columnwidth]{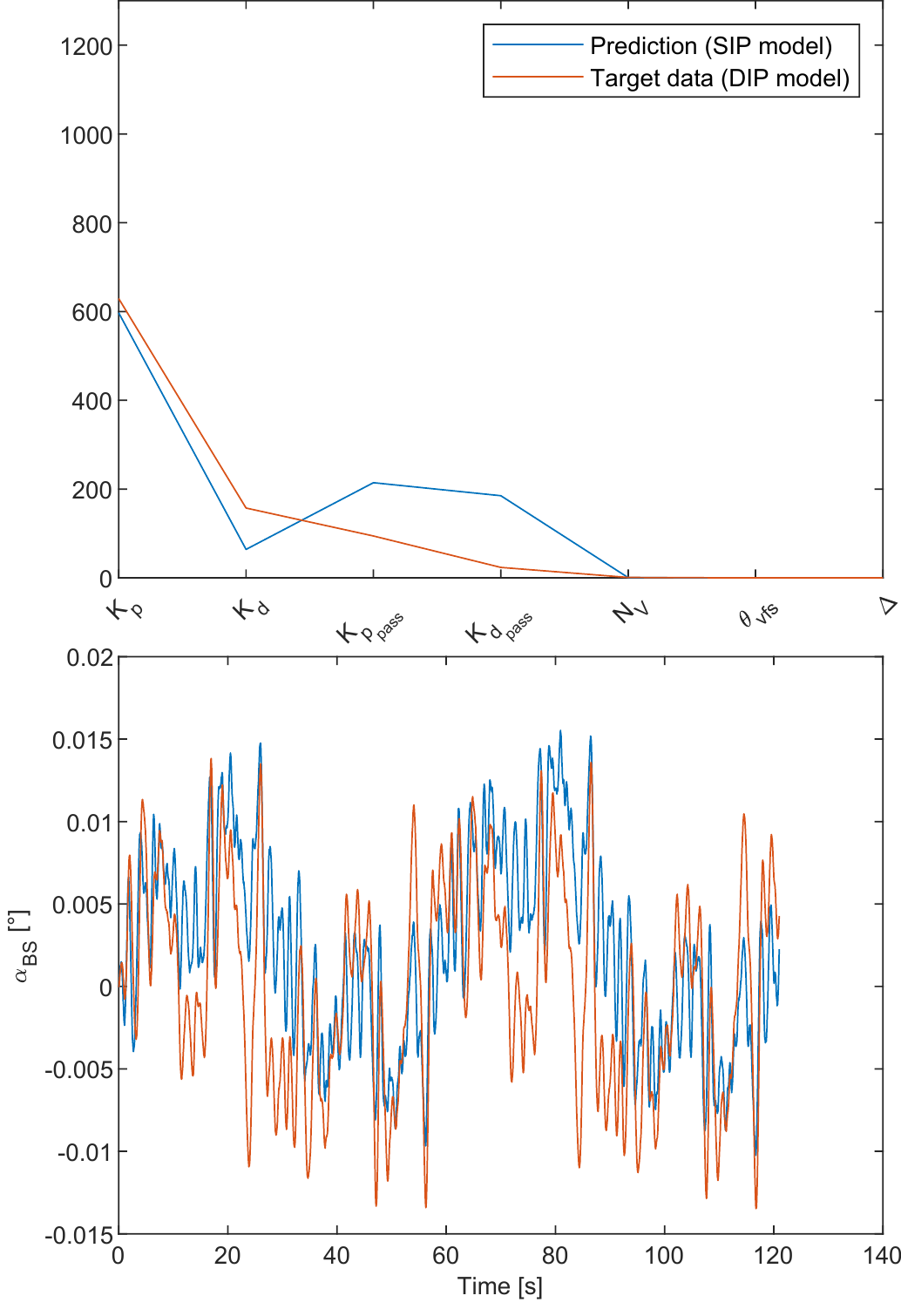}
		\caption{The CNN applied to a sample produced with a double inverted pendulum (DIP) model: (Top) network output compared with the parameters applied to the control of the ankle in the DIP model. The hip has also its set of parameters, not displayed. In (Bottom) the $\alpha_{BS}$ trajectory produced by the DIP is compared with the one of a SIP using the parameters predicted by the CNN.}
	\label{Dip2Sip}
\end{figure}

The system is now applied to DIP data produced with a simulation with default parameters\cite{Hettich2013}.
In Fig. \ref{Dip2Sip} the control parameters for the ankle module used in the simulation are compared with the ones identified by the CNN. The accuracy of the result is, as expected slightly worse than the one obtained using the validation set (with a SE (Squared Error) for the normalized parameters of $10.4783$ as compared to the smaller validation set MSE of $0.2851$). 
\subsection{Identification of Human posture control parameters}
\begin{figure}[htbp]
	\centering
		\includegraphics[width=1.00\columnwidth]{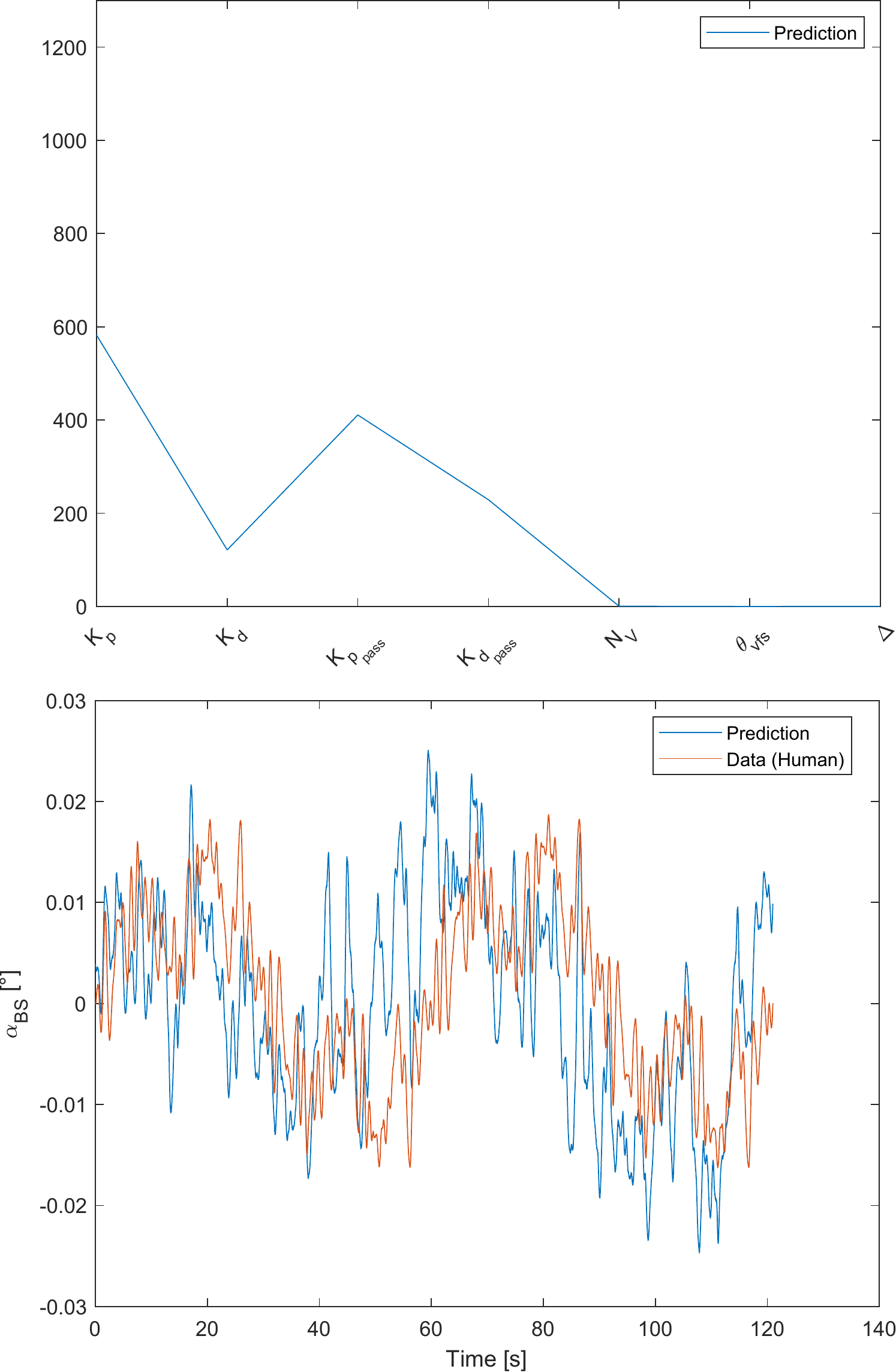}
	\caption{The CNN applied to data from a human experiment. (Top) The identified parameters are shown. (Bottom) The original input is compared with the input produced by the simulation using the parameters identified by the CNN.}
	\label{fig:ResultHuman}
\end{figure}

The CNN is here applied to human data. A single trial from one subject is used. Like in the previous example, the experiment does not include any device to block the hip so that the center of mass sway is influenced by the ankle-hip coordination. The identified parameters and the simulated body sway are shown in Fig. \ref{fig:ResultHuman}. The peak-to-peak sway amplitude exhibited by the human subject was $2.8533^{\circ}$. This example suggests that is beneficial to include larger body sway examples in the training-set. The result in Fig. \ref{fig:ResultHuman} shows a good, although not perfect, similarity between the simulation and the original data. 

\section{\uppercase{Conclusions and Future Work}}
\label{sec:conclusion}
In this work we presented a method for posture control parameter identification based on CNN. The system provides an efficient way to fit a model of the non-linear bio-inspired control system DEC on experimental data. This represents an advantage with respect to previous solutions relying on iterative methods. With the used training set noisy by design and because noise power is one of the parameters, the system provides an average error that is non-zero. The obtained parameters can be used a description of the analyzed data, as guideline for more accurate fitting procedures and as features for further analysis algorithms, for example a diagnostic tool could be implemented using SVM trained with the parameters vector as input.    

The training set is produced with parameters from uniform distributions, filtered with the constraint, checked empirically, so that they produce a stable simulation. In order to obtain more \textit{human-like} examples the data-set has been enriched with samples producing larger body sways. Future work may introduce preliminary studies on the distribution of human data, generating some training samples from sample obtained by fitting the model on the experimental data. The CNN can also be tested \textit{a posteriori} comparing the distribution of the parameters it produces on the validation set with the expected distribution on real data. This can help the process of choosing between different possible network hyperparameters sets as shown in \cite{sforza2011rejection,sforza2013support}.

The SIP model used in this work proved to be suitable to describe the analyzed posture control scenario, this even in the sub-optimal case of identifying the control parameter of the ankle joint in a DIP model. Future work will aim to the design of a solution that identifies also the parameters controlling the hip joint, which is known to have a relevant function in balancing \cite{G.Hettich2014,horak1986central,park2004postural}. 

The modeling and the analysis of human posture control and balance provides and get inspirations from the study of humanoid robots control, e.g. \cite{icinco07,icinco12,zebenay2015human,10.3389/fnbot.2018.00021}, or can be used to improve the design of assistive systems and devices \cite{icincoChugo.2019,Mergner2019}. The proposed deep-learning-based tool will be also published as a tool to benchmark humanoids and wearable devices \cite{torricelli2020benchmarking}, within the framework of the COMTEST project \cite{Lippi2019} that aims to make a posture control testbed available for the humanoid robotics community. 

\section*{\uppercase{Acknowledgements}}
\setlength{\intextsep}{-1pt}%
\setlength{\columnsep}{15pt}%
\begin{wrapfigure}{l}{0.08\columnwidth}
		{\includegraphics[width=0.12\columnwidth]{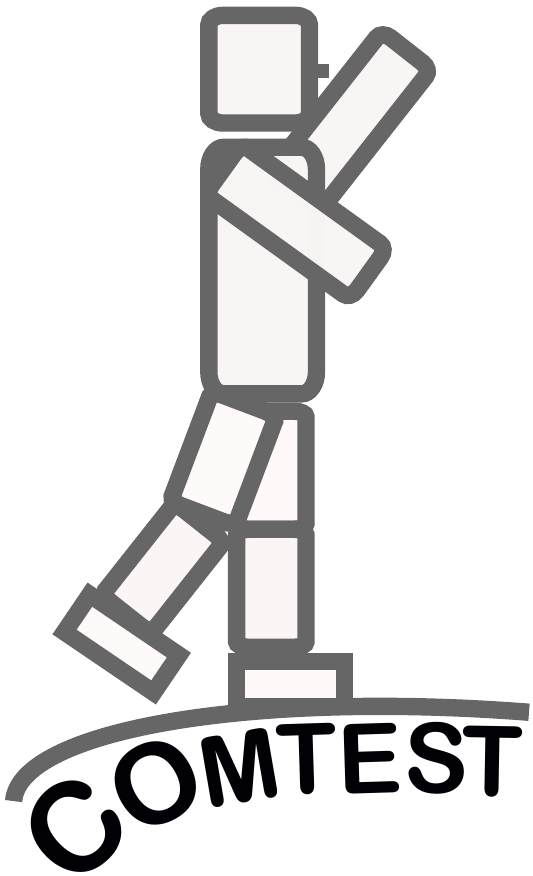}}
	\label{LOGO}
\end{wrapfigure}
\noindent This work is supported by the project COMTEST, a sub-project of EUROBENCH (European Robotic Framework for Bipedal Locomotion Benchmarking, www.eurobench2020.eu) funded by H2020 Topic ICT 27-2017 under grant agreement number 779963.

\bibliographystyle{apalike}
{\small
\bibliography{example}}

%

\end{document}